\documentclass[conference]{IEEEtran}
\IEEEoverridecommandlockouts
\usepackage{cite}
\usepackage{amsmath,amssymb,amsfonts}
\usepackage{graphicx}
\usepackage{textcomp}
\usepackage{xcolor}
\usepackage{url}
\usepackage{multirow}
\usepackage[hidelinks]{hyperref}

\usepackage{todonotes}
\usepackage{algorithm}
\usepackage{algorithmicx,algpseudocode}
\usepackage{subcaption}

\def\BibTeX{{\rm B\kern-.05em{\sc i\kern-.025em b}\kern-.08em
    T\kern-.1667em\lower.7ex\hbox{E}\kern-.125emX}}
\begin{document}
\title{Fully-fused Multi-Layer Perceptrons on Intel Data Center GPUs}

\author{%
  \IEEEauthorblockN{%
    Kai Yuan\IEEEauthorrefmark{2}\textsuperscript{*},
    Christoph Bauinger\IEEEauthorrefmark{2}\textsuperscript{*},
    Xiangyi Zhang\IEEEauthorrefmark{2}\textsuperscript{*},\\
    Pascal Baehr\IEEEauthorrefmark{2}, 
    Matthias Kirchhart\IEEEauthorrefmark{2},
    Darius Dabert\IEEEauthorrefmark{3},
    Adrien Tousnakhoff\IEEEauthorrefmark{3},   
    Pierre Boudier\IEEEauthorrefmark{2}
    and
    Michael Paulitsch\IEEEauthorrefmark{2}
  }
  \IEEEauthorblockA{\IEEEauthorrefmark{2} Intel Corporation}%
  \IEEEauthorblockA{\IEEEauthorrefmark{3} École Polytechnique}%
  
}

\maketitle

\begingroup\renewcommand\thefootnote{*}
\footnotetext{Equal contribution}
\begin{abstract}
This paper presents a SYCL implementation of Multi-Layer Perceptrons (MLPs), which targets and is optimized for the Intel Data Center GPU Max 1550. To increase the performance, our implementation minimizes the slow global memory accesses by maximizing the data reuse within the general register file and the shared local memory by fusing the operations in each layer of the MLP. We show with a simple roofline model that this results in a significant increase in the arithmetic intensity, leading to improved performance, especially for inference. We compare our approach to a similar CUDA implementation for MLPs and show that our implementation on the Intel Data Center GPU outperforms the CUDA implementation on Nvidia's H100 GPU by a factor up to 2.84 in inference and 1.75 in training. The paper also showcases the efficiency of our SYCL implementation in three significant areas: Image Compression, Neural Radiance Fields, and Physics-Informed Machine Learning. In all cases, our implementation outperforms the off-the-shelf Intel Extension for PyTorch (IPEX) implementation on the same Intel GPU by up to a factor of 30 and the CUDA PyTorch version on Nvidia's H100 GPU by up to a factor 19. The code can be found at \url{https://github.com/intel/tiny-dpcpp-nn}.
\end{abstract}


\section{Introduction}

Multi-Layer Perceptrons (MLPs) \cite{yegnanarayana2009artificial} play a vital role in today's Machine Learning (ML) and Artificial Intelligence (AI) landscape next to the prevalent Transformer architecture~\cite{vaswani2017attention} and Convolutional Neural Networks~\cite{li2021survey}, which are mostly used in the Natural Language Processing and Computer Vision domains. MLPs are used as the main Neural Network architecture for several Machine Learning applications, such as the representation of the solution operator of partial differential equations~\cite{cuomo2022scientific}, the density or color function in Neural Radiance Fields (NeRFs) objects~\cite{mildenhall2021nerf}, and replacing classical ray-tracing with Neural Ray Tracing~\cite{muller2021real} (see Section~\ref{sec:apps} for details). In contrast to the aforementioned architectures, MLPs are characterized by their fully connected layers, where the neuron of every layer is connected to every neuron in the previous and next layer. A key distinguishing factor is that in MLPs, each neuron's output is independent of its neighbors in the same layer, making it suitable for fully-fusing operations as described in this work.

The present contribution focuses on the efficient implementation on Intel GPUs of ``narrow'' MLPs, which consist of an arbitrary number of layers (depth), and a small and constant number of neurons per layer (width). These narrow MLPs are of particular interest since i) they are universal approximators~\cite{park2020minimum}, ii) they have several relevant use cases (see Sec.~\ref{sec:apps}), and iii) their theoretical peak performance is severely limited by the small width of the layers. More precisely, their small width results in a reduced arithmetic intensity of the requisite matrix-multiplications in each layer for the training and, in particular, the inference. Thus, in a ``classic'' implementation of MLPs, where necessary operations in each layer are performed in separate compute kernels, the performance of narrow MLPs is typically severely bound by the memory bandwidth of the global memory or the last level cache. 

To alleviate the issues arising from the low arithmetic intensity and the memory bandwidth of the global memory, a common strategy~\cite{muller2021real} is the fusion of the layers into a single kernel to keep relevant data in faster memories, i.e., the register file, shared memory or faster caches. This approach, termed ``fully-fused MLPs'', has been implemented for Nvidia GPUs utilizing Nvidia's CUDA language~\cite{tiny-cuda-nn}. 

In this contribution we focus on a SYCL implementation for Intel GPUs of fully-fused MLPs with arbitrary depth and fixed layer width of $2^i, i \in \{4,..., 7\}$ neurons in each layer. Note that, as indicated above, fixed widths are not limiting the expressive power of the Neural Network as fixed-width MLPs are still universal approximators, i.e., they can approximate any continuous function to any desired accuracy as proven by the Universal Approximation Theory for width-bounded networks~\cite{park2020minimum}. 
Indeed, in practise MLPs rarely exceed the maximum width of $2^7$ elements supported by this work as networks tend to be deeper to gain more expressiveness rather than wide (see Section \ref{sec:apps}).

Our implementation of fully-fused MLPs is based on Intel's joint\_matrix SYCL extension~\cite{intel_optimization_guide} to utilize the XMX hardware~\cite{peddie2023sixth} in Intel's Data Center GPU Max 1550~\cite{jiang2022intel}, which is the device targeted with our optimized implementation. 

Our method is especially well-suited to optimize the training and inference performance for models that require large data throughput with batch sizes $2^i, 15 < i \in \mathbb{N}$, since those sizes maximize the occupancy of the device. As shown in Section~~\ref{sec:benchmark}, our SYCL implementation on Intel hardware has improved performance over an equivalent CUDA implementation (tiny-cuda-nn~\cite{tiny-cuda-nn}) for MLPs with width 64 by a factor up to 2.84 in inference and 1.75 in training.

Furthermore, we argue with a roofline analysis (see Section~\ref{sec:roofline}) that our approach to fully-fused MLPs is especially well-suited for the acceleration of the inference and that it significantly increases the arithmetic intensity, and thus the theoretical peak performance, compared to the approach shown in~\cite{muller2021real} by reducing the accesses to global memory.

To further show the performance improvements and potential applications for our implementation, we demonstrate the performance on a regression benchmark and the following three applications: Image Compression, Neural Radiance Fields (NeRFs), and Physics-Informed Machine Learning (see Section \ref{sec:results}).  In all cases, our implementation outperforms the off-the-shelf Intel Extension for PyTorch (IPEX) \cite{ipex} implementation on the same Intel GPU by up to a factor of 30 and the CUDA PyTorch version on Nvidia's H100 GPU~\cite{h100} by up to a factor 19. 

\textbf{Summarised, the contributions of this paper are:}
\begin{itemize}
    \item[1.] The first \href{https://github.com/intel/tiny-dpcpp-nn}{SYCL implementation of fully-fused MLPs} applied on Intel GPU's that support XMX instructions and open-sourced repository of the implementation.
    \item[2.] A roofline model of our implementation and comparison to the roofline of the fully-fused implementation~\cite{muller2021real}. We argue an improvement of the arithmetic intensity of up to a factor 2.15.
    \item[3.] Demonstrating higher performance on four example applications: regression benchmark, image compression, Neural Radiance Fields, and Physics-Informed Neural Networks. Demonstrating an improvement of up to 1.75x and 2.84x performance increase for training and inference respectively over another fully-fused implementation, and a performance increase of a factor up to 30 over off-the-shelf PyTorch implementations.
\end{itemize}

The following sections are structured as follows. First, we outline the applications of MLPs and their impact. Next, the fully-fused MLPs are described. Finally, we demonstrate our results on four example applications, and conclude this paper.

\section{Applications of Multi-Layer Perceptrons}
\label{sec:apps}

To show the importance of MLPs in today's AI landscape, this section reviews the applications of MLP variants and implementations in multiple areas. Please note, that due to the fast-paced adoption of AI and ML in various fields and industries, this list is non-exhaustive and should rather provide the reader with a qualitative intuition about the importance of accelerating MLPs and their impact to the community. For a more comprehensive overview, please refer to ~\cite{liu2022we, delashmit2005recent, guo2021can}.

Further, to provide a better understanding of what performance gains can be achieved in key areas, we highlight and provide more details on three prominent applications in the Appendix~\ref{sec:nerf_explanation}-\ref{sec:piml_explanation}: Neural Rendering, Model Compression, and Partial Differential Equations. The performance and results generated with our implementation for these applications are shown in Section~\ref{sec:results}. 

\begin{figure}
    \centering
    \includegraphics[width=1\linewidth]{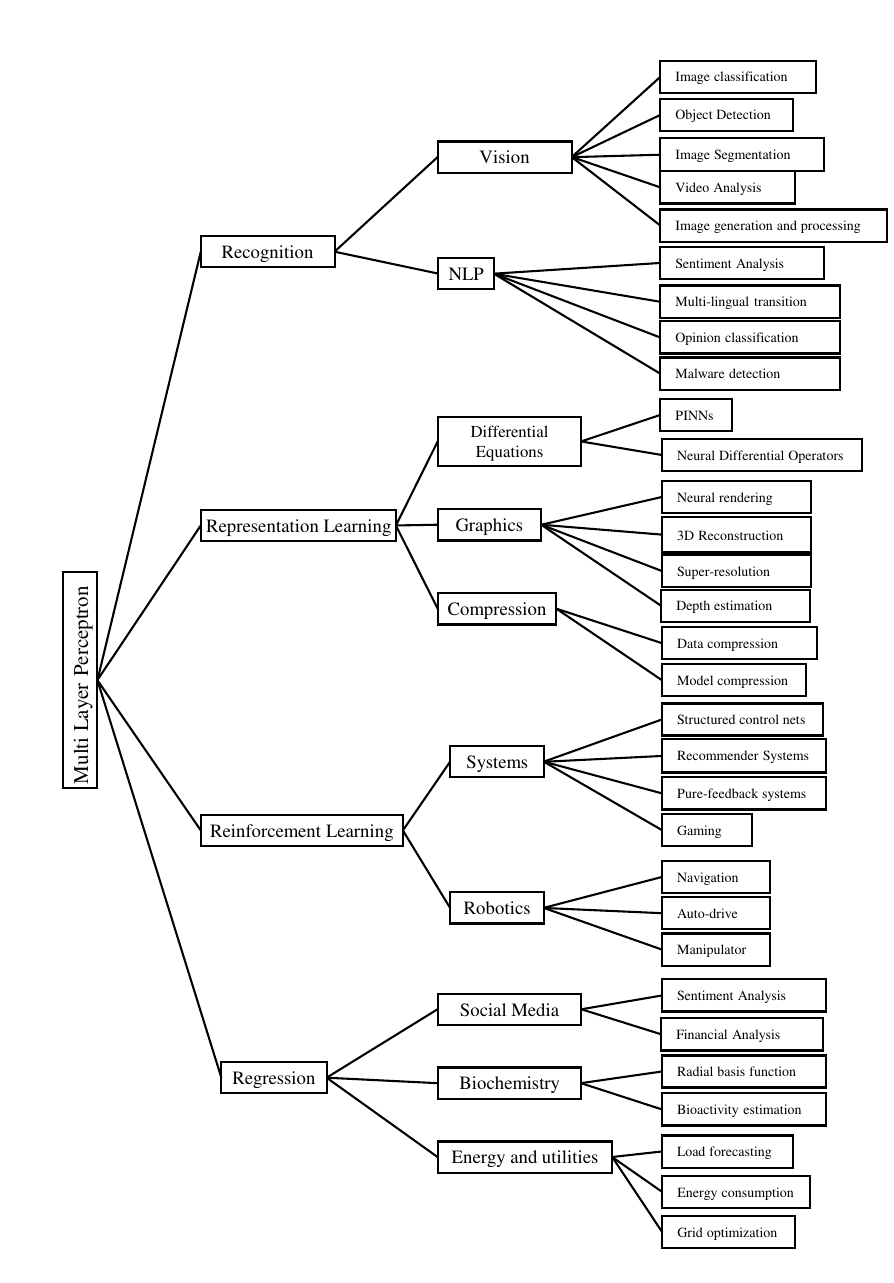}
    \caption{Taxonomy categorizing various applications of MLP. Each category has multiple subcategories that illustrate the specific tasks and domains that MLP can address.}
    \label{fig:taxonomy} 
\end{figure}


Most MLP applications can be categorized into the following four categories: 
\begin{itemize}
    \item Recognition: using MLPs to identify  objects, or patterns.
    \item Representation learning: aiming to extract meaningful patterns from raw data to create representations that are easier to understand and process by MLPs.
    \item Reinforcement learning: utilizing MLPs to learn from interactions with the environment  to make better decisions. 
    \item Regression: predicting values based on input data.
\end{itemize}

The taxonomy in Figure \ref{fig:taxonomy} categorizes contemporary MLP applications, which demonstrates the diversity and versatility of MLP in solving various real-world problems. We present these categories in the following sections in some detail.

\subsection{Recognition}
In vision, MLPs are applied to image classification~\cite{hou2021vision,ding2022repmlp,rao2021global,touvron2021resmlp,liu2021pay}, objects detection and semantic segmentation~\cite{chen2022cyclemlp, lahoud2017object, pan2022unext, an2021abdomen, lai2022axial}, video analysis~\cite{qiu2022mlp,an2021morphmlp, xia2022skating}, image processing and generation~\cite{tu2022maxim,sharma2017new,cazenavette2021mixergan,mansour2022image}.
Other notable works in NLP include sentiment analysis and opinion classification
~\cite{al2020improving, sultana2018prediction, al2019automatic}, malware classification~\cite{tran2017nlp, dang2021malware, qiao2021malware}, adversarially robust~\cite{raman2023model} and multilingual transition~\cite{fusco2022pnlp} which leverages MLPs for language understanding and generation.

\subsection{Representation Learning}

MLPs may solve partial differential equations through Physics-Informed Neural Networks (PINNs)~\cite{alkhalifahphysics, takamoto2023learning, eivazi2022physics, cuomo2022scientific, wang2022respecting} and neural differential operators~\cite{lu2019deeponet, he2023neural, zhang2019multilayer}, which  approximate the solution to differential equations using neural networks.

In the field of neural graphics, MLPs have been developed
and integrated into neural rendering~\cite{mildenhall2021nerf, garbin2021fastnerf, barron2021mip,reiser2021kilonerf, sun2022direct,park2021nerfies}, 3D reconstruction~\cite{cao2022jiff, zhu2023reconstruction,bozic2021neural}, super-resolution~\cite{peyrard2014comparison} and depth-estimation~\cite{lee2017estimating, chopade2019single}. In this work, we are interested in exploiting the potential of MLPs in the above three fields and will show their results in the  Appendix~\ref{sec:nerf_explanation}-\ref{sec:piml_explanation} for details.

In the field of Compression, MLPs aim to reduce the size of data or models by removing redundant or irrelevant parameters and operations. MLPs are utilized to perform data compression by learning compact representations of data that can be reconstructed with minimal distortion\cite{juszczyk2016application, strumpler2021implicit}. In addition, MLPs can be utilized for model compression by learning to approximate the function of large or complex models with smaller or simpler models~\cite{zhao2022analysis,yin2021tt}. 

\subsection{Reinforcement Learning}

In the field of robotics, MLPs are integral for various tasks, including navigation~\cite{liu2016path, jalali2020neuroevolution, ko2017neural, singh2018mobile}, autonomous-driving~\cite{zhang2019autonomous}, and manipulator~\cite{song2021pc}.
In system research, MLPs have been employed for structured control nets~\cite{song2021pc, srouji2018structured}, recommender systems~\cite{chen2023deep,zhao2020ameir}, pure-feedback systems~\cite{liu2019mlp, bai2020nn} and Gaming~\cite{bjorck2021towards, wagenaar2017learning}. 
which utilize MLP-based deep learning techniques to discover system design and control.

\subsection{Regression}
\label{sec:regression}
Most existing works using MLPs on regression are in the fields of social media, biochemistry, and Energy. 

In social media, MLPs play a pivotal role in sentiment analysis~\cite{carosia2020analyzing, bairavel2020novel, sun2022cubemlp, ramdhani2022sentiment}, financial analysis~\cite{abedin2019topological, neagoe2018deep} and fraud detection\cite{pehlivanli2019detection, makuyana2020fraud}, for analysis the social problems.
Furthermore, in biochemistry research, MLPs are able to model complex biology and chemistry to explore the molecular basis of life in radial basis function~\cite{cheshmberah2020comparison} and bioactivity estimation~\cite{liu2016mlp}.
In the field of energy and utilities, MLPs work well on load forecasting~\cite{park2016approximate}, energy consumption~\cite{taki2018assessment}, and machine optimization~\cite{esmaeili2014multiple} to deal with some biological issues and improve the effect of activities.

\section{Fully-fused MLPs on Intel GPUs}
In the present section, we introduce our approach to the implementation of fully-fused MLPs on Intel GPUs. Section~\ref{sec:operations} we describe the inference and training algorithms. Section~\ref{sec:implementation} then describes our SYCL joint\_matrix implementation for the Intel Data Center GPU Max 1550 in detail.

\subsection{Inference and Training Operations} 
\label{sec:operations}

For the sake of conciseness and simplicity, we describe our implementation of fully-fused MLPs for the special case where the width of each layer, including input and output, is 64. Cases where the input and output size differs may be reduced to this special case by utilizing, e.g., {\em encodings} for the input data or appropriate padding of the data and the weights. 

The matrix operations  during training and inference are described in what follows.
Let $M\in \mathbb{N}$ denote the batch size and let $N=K=64$ denote the layer width.
Let $\sigma : \mathbb{R} \to \mathbb{R}$ denote an activation function~\cite{sharma2017activation}. We use the notation $\sigma(A)$, for a matrix $A$, to indicate the element-wise application of the activation function to the matrix $A$. Let $2 \leq \mathrm{nlayers} \in \mathbb{N}$ denote the number of layers in the MLP, which includes the input and the output layer. I.e., it denotes the number of matrices as in Alg.~\ref{alg:inference}. Further, let $A_i \in \mathbb{R}^{M \times K}$, $B_i \in \mathbb{R}^{K \times N}$, and $C_i \in \mathbb{R}^{M \times N}$, $D_i \in \mathbb{R}^{M \times N}$, $G_i \in \mathbb{R}^{K \times N}$ denote matrices for each $i=1, \ldots, \mathrm{nlayers}$. The transpose of any matrix $A$ is given by $A^T$. An element of a matrix $A$ with the row coordinate $r$ and the column coordinate $c$ is given by $A(r,c) \in \mathbb{R}$. The matrices $B_i$ denote weights of each layer $i$, $1 \leq i \leq \mathrm{nlayers}-1$.

A pseudo code of the inference is shown in Algorithm~\ref{alg:inference}, which consists of repeated matrix multiplications $A_i B_i$ and applications of the activation function $\sigma$. 
\begin{algorithm}[ht]
\small
\caption{Inference}
\begin{algorithmic}
\Require $\mathrm{nlayers}$, $\sigma$, $\mathrm{Input}$, $B_1, \ldots, B_{\mathrm{nlayers}-1}$
\State {Initialize: $A_1 = \mathrm{Input}$ }
\For {$ i \leftarrow 1$ \textbf{to} $\mathrm{nlayers}-1$}
\State {$C_i \leftarrow A_i B_i$}
\State {$A_{i+1} \leftarrow \sigma(C_i)$}
\EndFor
\State {Return $A_{\mathrm{nlayers}}$}
\end{algorithmic}
\label{alg:inference}
\end{algorithm}

Our implementation of the inference step is based on several observations. First, each weight matrix $B_i$ fits in the shared local memory (SLM)~\cite{slm_optguide} of the targeted Intel Data Center GPU Max 1550. The loads of the $B_i$ matrices from global memory, i.e., high bandwidth memory (HBM), can therefore be minimized by pre-fetching them one after the other into SLM. Secondly, up to eight rows of the matrices $A_i$ and $C_i$ (for a single $i$) fit into the general register file (GRF) for all network widths which are 64 or less. For network widths between 64 and 128, it still holds but requires large GRF mode~\cite{grfmode}. This is relevant since the targeted XMX hardware may be applied to sub-matrices with up to eight rows. However, as analyzed in the following Sec.~\ref{sec:roofline}, the highest arithmetic intensity is achieved when using sub-matrices consisting of exactly eight rows. Lastly, only $A_i$ and $C_i$ for a single $i$ are required at the same time and only the last matrix $A_{\mathrm{nlayers}}$ is returned. The matrices associated with other layers $j$,  $1 \leq j \leq \mathrm{nlayers}-1$ are discarded during inference thus minimizing the amount of memory accesses required, consequently increasing the arithmetic intensity and performance.

The pseudo code for the training of the network is shown in Algorithm~\ref{alg:training}. In particular, we focus on the forward pass and the subsequent backward pass after a loss is calculated. We do not consider the optimization step in this pseudo code.
In contrast to the inference, the training takes as input two activation functions. One for the forward pass and one for the backward pass denoted by $\sigma_f$ and $\sigma_b$, respectively. Note, that the activation function $\sigma_b$ is the derivative of $\sigma_f$ resulting from the chain rule during back-propagation \cite{goodfellow2016deep}. In addition, the loss computation requires target values given as an $M \times K$ array of real values, named $\mathrm{Target}$. The outputs of the training algorithm are $\mathrm{nlayers}-1$ matrices $G_1, \ldots, G_{\mathrm{nlayers}-1}$. The matrices $G_i$ represent the gradients of the weight matrices $B_i$. The gradients are required for the subsequent optimization step, which is typically some type of gradient descent algorithm and is a real square matrix of size $K \times N$ (note that $K=N$ as we have fixed layer widths of $2^i, i = 4,..., 7$) .

The forward pass in Algorithm~\ref{alg:training} is nearly identical to the inference. The only difference is that all matrices $A_i$, $i=1, \ldots, \mathrm{nlayers}$, have to be stored since they are required for the backward pass. The loss calculation in Algorithm~\ref{alg:training} is an exemplary L2 loss. Note, that the loss calculation is explicitly mentioned as $\mathrm{Loss}(\mathrm{row}, \mathrm{col})$ for illustration purposes but not stored as the derivative of the loss is used for the back-propagation step (see step i) in the following paragraph).

The backward pass consists of three steps: i) propagating the gradient backward to the next layer, ii) calculate the gradient of the loss with respect to the weighted inputs, and iii) calculate the gradient of the loss with respect to the weights. As follows from the chain rule for back-propagation, i) in each layer $i$ the gradient propagation is realized as matrix multiplications between the matrix $D_{i+1}$ and the transpose of the weights matrix $B_i$, ii) gradients with respect to weighted inputs are calculated as activation of the previous layer, and iii) gradients with respect to the weights are a matrix multiplication between the transpose of the forward outputs $A_i^T$ and the activated output of the backward pass. For a  derivation of the backward pass, please refer to Chapter 6.5 in \cite{goodfellow2016deep}.

Our implementation of the training computes the forward pass, the loss and the backward pass within the same kernel to maximize the re-use of cached data and registers. Only the product $A_{i-1}^T D_i$ is not part of our training kernel, although the time required for it is included in all of the performance data presented in the following Section~\ref{sec:results}. We found that it is in general preferable to launch a separate kernel for the products $A_{i-1}^T D_i$ due to the differing dimensions of the matrix multiplications. In detail, a possible version which fuses the matrix multiplications $A_{i-1}^T D_i$ into the training and which reuses the data already in the GRF would perform a block outer product of a block row in the matrix $A_i$ with a block row at the same position in the matrix $D_i$ to compute a partial sum of the whole $K \times N$ output matrix. A subsequent reduction over all work-items would then be necessary for the final result. At the time of writing, we were not able to find an efficient implementation of the above approach or any other approach to the fusing of the product $A_{i-1}^T D_i$. We instead found that performing the operations $A_{i-1}^T D_i$ with the highly optimized matrix-multiplication routines in Intel's oneMKL, in a separate kernel after our training implementation finishes, delivers the best results.

\begin{algorithm}[ht]
\small
\caption{Training}
\begin{algorithmic}
\Require $\mathrm{nlayers}$, $\sigma_f$, $\sigma_b$, $\mathrm{Input}$, $\mathrm{Target}$, $B_1, \ldots, B_{\mathrm{nlayers}-1}$
\State {Initialize: $A_1 = \mathrm{Input}$ }
\For {$ i \leftarrow 1$ \textbf{to} $\mathrm{nlayers}-1$} \Comment{Forward Pass}
\State {$C_i \leftarrow A_i B_i$}
\State {$A_{i+1} \leftarrow \sigma_f(C_i)$}
\EndFor
\For {$\mathrm{row} \leftarrow 1,\ldots,M$} \Comment{Loss Calculation}
\For {$\mathrm{col} \leftarrow 1,\ldots,K$}
\State {$\mathrm{Diff}(\mathrm{row},\mathrm{col}) = A_\mathrm{nlayers}(\mathrm{row}, \mathrm{col}) - \mathrm{Target}(\mathrm{row},\mathrm{col})$}
\State{ $\mathrm{Loss}(\mathrm{row}, \mathrm{col}) = \mathrm{Diff}(\mathrm{row},\mathrm{col})^2 / (M K)$ }
\State{$D_\mathrm{nlayers}(\mathrm{row}, \mathrm{col}) = 2\mathrm{Diff}(\mathrm{row},\mathrm{col})/ (M K)$}
\EndFor
\EndFor
\State {$D_{nlayers} \leftarrow \sigma_b(A_{nlayers}, D_{nlayers})$}
\State {$G_{nlayers-1} \leftarrow A_{nlayers-1}^T D_{nlayers}$}
\For {$i \leftarrow \mathrm{nlayers-1}$ \textbf{to} $2$} \Comment{Backward Pass}
\State {$D_i \leftarrow D_{i+1} B_{i}^T$}
\State {$D_i \leftarrow \sigma_b(A_i, D_i)$}
\State {$G_{i-1} \leftarrow A_{i-1}^T D_{i}$}
\EndFor
\State {Return $G_1, \ldots, G_\mathrm{nlayers-1}$}
\end{algorithmic}
\label{alg:training}
\end{algorithm}

\subsection{SYCL joint\_matrix Implementation of MLPs} \label{sec:implementation}
Our SYCL implementation is based on Intel's \textit{joint\_matrix} extension~\cite{intel_optimization_guide} to write code which can be executed on Intel's XMX hardware~\cite{xe_arch} for the matrix multiplications. Efficient utilization of the XMX hardware is highly relevant since the theoretical multiply-add (MAD) peak performance of an Intel Data Center GPU Max 1550 on the targeted bfloat16~\cite{bfloat16} (bf16) data type is approximately 838 tera floating point operations per second (Tflops/s), which is sixteen times the theoretical peak single-precision MAD throughput ($\sim 52$ Tflops/s) when utilizing the vector engines~\cite{xe_arch}.

The Intel SYCL extension provides a joint\_matrix object, which represents a matrix of a small and fixed size distributed across a SYCL sub-group~\cite{intel_optimization_guide}, and several related functions to facilitate computations with these joint\_matrix objects. In particular, we utilize the function joint\_matrix\_mad, which takes three joint matrices, say $A$, $B$, and $C$, and returns the joint\_matrix resulting from the operation $AB + C$, which is performed on the XMX hardware. 

The joint\_matrix\_mad function is only available for specific matrix sizes. In detail, the matrix $A$ has to be of size $\mathrm{TM}\times \mathrm{TK}$ where $\mathrm{TM} \in \{1, \ldots, 8\}$ can be chosen arbitrarily and $\mathrm{TK}$ depends on the device and the data type~\cite{joint_matrix_extension_github2}. For the targeted bfloat16 data type, $\mathrm{TK} = 16$ is required. The matrix $B$, in turn, has to be of size $\mathrm{TK} \times \mathrm{TN}$, where $\mathrm{TN}$ is device dependent. For the Intel Data Center GPU Max the value $\mathrm{TN} = 16$ is required. The matrix $C$ as well as the output matrix of the joint\_matrix\_mad function have to be of size $\mathrm{TM} \times \mathrm{TN}$. It is important to note that joint\_matrix\_mad performs the accumulation in single precision when utilizing bfloat16 for the inputs $A$ and $B$. The matrix $C$ and the output of the  function are therefore of type float.

In the following detailed description, we focus on the inference (cf. Alg.~\ref{alg:inference}) for the special case where $K=N=64$. We launch the kernel with $M / \mathrm{TM} \times 16$ work-items, a sub-group size of $16$ and a work-group size of $1024$ work-items, i.e., the maximum possible work-group size to minimize the loads from HBM as discussed in the following Section~\ref{sec:roofline}. Each sub-group with id $\mathrm{sg\_id}$, $\mathrm{sg\_id} = 1, \ldots, M/\mathrm{TM}$ loads a unique block-row from the input into the register file starting at the row $\mathrm{TM}\times\mathrm{sg\_id}$ consisting of $\mathrm{TM}$ rows and $K$ columns. Each sub-group stores this block-row as four joint\_matrix objects, each of size $\mathrm{TM}\times \mathrm{TK}$. In each layer $i$, each work-group (consisting of 64 sub-groups) loads the weight matrix $B_i$ jointly from HBM into SLM. At the end of the algorithm, each sub-group stores its $\mathrm{TM}$ rows of the matrix $A_{\mathrm{nlayers}}$ to HBM.

Figure~\ref{fig:matrix_mult} illustrates the loads to the register file (green) per sub-group and the joint load to SLM (blue) per work-group for a single layer $i$. Each sub-group then computes the product of its block-row with the weights matrix $B_i$ and keeps the resulting four joint\_matrix objects ($K=64$, $\mathrm{TK}=16$) in the registers. The weight sub-matrices sized $\mathrm{TK} \times \mathrm{TN}$, which are required to facilitate the joint\_matrix\_mad function, are loaded on-the-fly from SLM as needed using the joint\_matrix\_load~\cite{joint_matrix_load_docu} function. The weights are stored in a packed format~\cite{joint_matrix_extension_github} to increase the performance of the load.

After the matrix-matrix product, each sub-group applies the activation function $\sigma$ to its resulting block-row and utilizes the output of the activation function in the next iteration as input to the subsequent matrix-matrix product. Thus, it keeps the data in the register file and avoids accesses to HBM or caches. This idea is similar to ~\cite{muller2021real}. However, our approach keeps the weights matrix in SLM and the input $A_i$ and output $C_i$ in the registers. Our algorithm requires two local synchronizations per layer due to the utilization of the SLM for the weights. The first synchronization ensures that the copy of the weight matrix into SLM is finished before accessing it. The second synchronization is to ensure that every work-item in the work-group finished accessing the weight matrix in SLM before starting the copy the weight matrix for the next layer.

The training algorithm~\ref{alg:training} is similar to the inference described above and follows the same structure. Although, in contrast to the inference, each layer $i$ in the training has to store the matrices $A_i$ and $D_i$, which may impact the performance. As indicated above, the final matrix-matrix multiplications, $A_i^T D_i$, are performed by oneMKL outside of our training kernel since no fusing is possible for these operations.

\begin{figure}
    \centering
    \includegraphics[width=\linewidth]{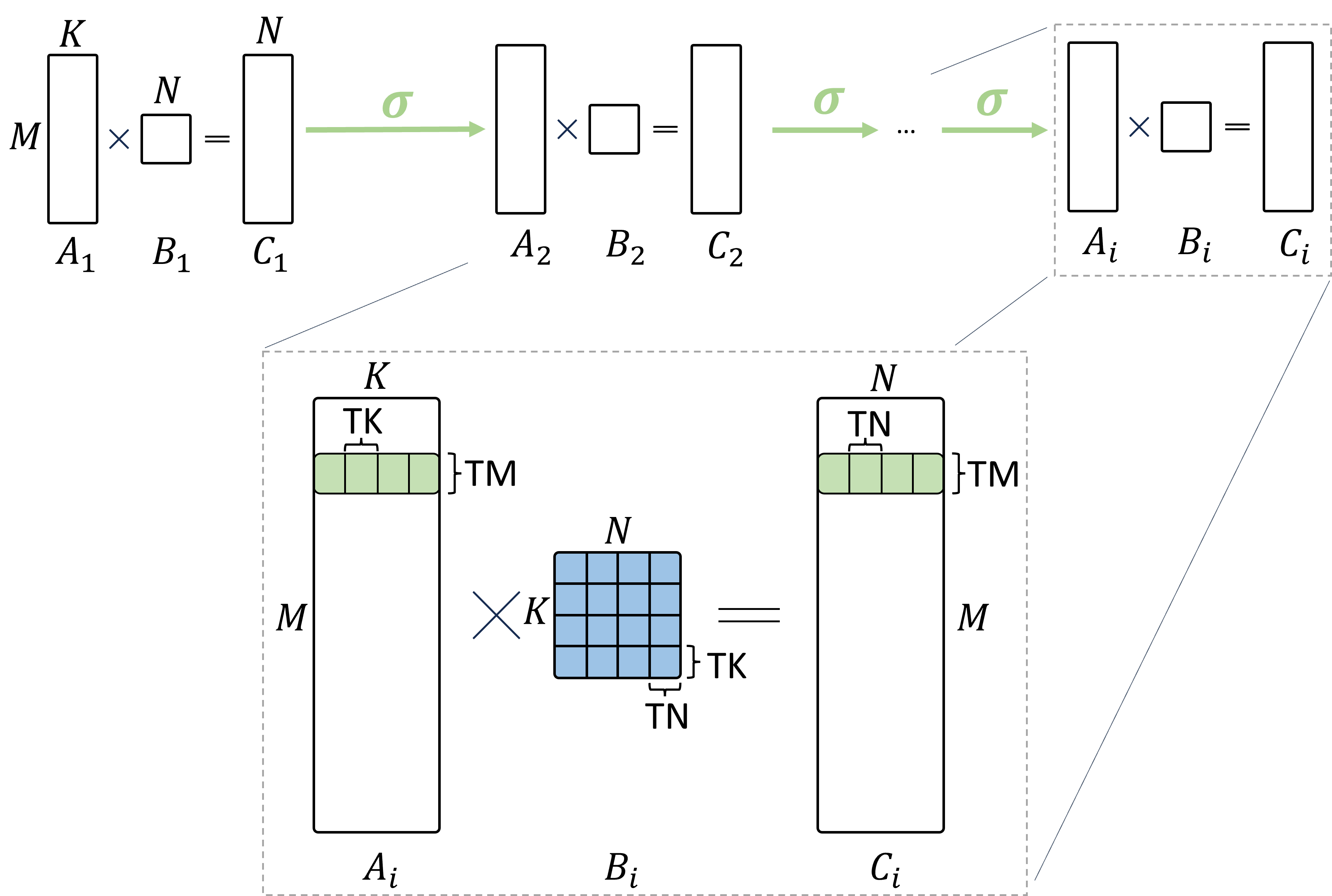}
    \caption{Illustration of our implementation. On the left hand-side, all multiple layers are sketched. Each layer is parallelized along the batch size (i.e., the $M$ dimension). The right hand-side sketches a single layer. The data of a single sub-group is colored. Green indicates data in the register file. Blue indicates data in the SLM. Each sub-group performs a matrix-matrix multiplication of size $(\mathrm{TM}\times\mathrm{K}) \times (\mathrm{K}\times\mathrm{N})$ utilizing joint\_matrix objects of size $\mathrm{TM}\times \mathrm{TK}$, $\mathrm{TK}\times \mathrm{TN}$, and $\mathrm{TM}\times \mathrm{TN}$.}
    \label{fig:matrix_mult}
\end{figure}

To conclude this section, we discuss the limitations of our approach and propose mitigation strategies.

First, to achieve an occupancy of 100\% on the Intel Data Center GPU Max 1550, 8192 sub-groups have to be launched. Thus, problems with a batch size of less than $8192\times8=65536=2^{16}$ ($\mathrm{TM} = 8$) do not fully occupy the device with our strategy and show reduced performance. This restriction is alleviated by choosing a smaller value for $\mathrm{TM}$, which increases the number of launched sub-groups. As discussed in Section~\ref{sec:roofline} this reduces the arithmetic intensity and may lead to a higher occupancy at the cost of a lower arithmetic intensity.

Secondly, due to the fixed dimensions of the inputs to the joint\_matrix\_mad functions, the batch size $M$ has to be a multiple of $\mathrm{TM}$ and the network width $N=K$ has to be a multiple of $\mathrm{TN}$ and $\mathrm{TK}$. This limits the possible width of the network to multiples of 16 on the Intel Data Center GPU Max 1550. These limitations can be removed by simply padding the batch size to the next multiple of $\mathrm{TM}$ and the network width to the next multiple of 16. Another alternative would be to use two-dimensional load functions and hardware-supported zero-padding of out-of-bounds accesses for these two-dimensional loads. Although, at the time of writing, support of two-dimensional loads is currently lacking in Intel's SYCL implementation and may required to use intrinsics and built-in functionalities. 

Finally, our algorithm is based on storing whole block-rows in the register file. The maximum possible width of the network is therefore determined by the size of the register file and given as 128 elements when utilizing the large register mode~\cite{grfmode}. In fact, assuming that the sizes of the data types of the inputs, the weights, and the output of each layer are given as, say, $s_A$, $s_B$, and $s_C$ bytes, respectively, we allocate $\mathrm{TM}\times K \times s_A + \mathrm{TM}\times N \times s_C + \mathrm{TK}\times\mathrm{TN}\times s_B$ bytes per sub-group in the register file at the same time to facilitate the joint\_matrix multiplications (3584 bytes for $N=K=64$, $\mathrm{TM} = 8$, $\mathrm{TK} = \mathrm{TN} = 16$, $s_A = s_B = 2$, and $s_C = 4$.) While the default size of the register file of the Intel Data Center GPU Max 1550 is 8192 bytes per sub-group, i.e., more than twice the size required for the joint\_matrix multiplications, we empirically found that a width larger than 64 elements does not fit in the GRF and thus results in register spills. Although, the register allocation depends on the driver version and may change in the future. Since the large register mode doubles the GRF size, MLPs with a width of 128 elements fit. The register pressure may be reduced by choosing a smaller value for $\mathrm{TM}$ at the cost of performance (see following Section~\ref{sec:roofline} for the relation between $\mathrm{TM}$ and the arithmetic intensity).

\subsection{Roofline Analysis}
\label{sec:roofline}
Similar to~\cite{muller2021real}, our goal is to minimize the accesses to the relatively slow HBM. To analyze how well our implementation achieves this goal, we compare our implementation to the CUDA implementation in~\cite{muller2021real} utilizing a roofline model~\cite{roofline_model} with bandwidth limits. We compute and compare the arithmetic intensities of the algorithms for $\mathrm{TM} = 8$, $K=N=64$ and the bfloat16 data type (i.e., 2 byte per value in memory). A higher arithmetic intensity indicates that more floating point operations per byte can be performed and thus shows that the performance is less limited by the memory bandwidth. Note, that a higher arithmetic intensity does not necessarily translate to a higher throughput since other factors (e.g. latency, frequency throttling, etc.) may be limiting the performance. 
Further note, that we are not considering the floating point operations performed by the activation function in the following analysis. 

 We use 1024 work-items per work-group and a sub-group size of 16 work-items. Thus 64 sub-groups (1024/16) constitute one work-group. The number of loads of the weights matrix from HBM is inversely proportional to the number of sub-groups in a work-group (see Section~\ref{sec:implementation}). As each sub-group computes $\mathrm{TM}$ rows of the $M$ rows of the input, we launch $M/\mathrm{TM}$ sub-groups or $M/\mathrm{TM}/64 = M / \mathrm{TM} \times 16 / 1024$ work-groups. Each work-group has its own SLM and thus each work-group needs to load the weights matrix from HBM into its SLM. Note, that in our particular case the weights matrix may be cached in L2. 
 We neglect the effects of the L2 cache in our roofline analysis, as it would depend on too many parameters, including $N$, $K$, $M$, $\mathrm{TM}$, $\mathrm{nlayers}$, and datatypes.

As a consequence, for a single layer $i$, $i=1,\ldots,\mathrm{nlayers}-1$, under the above mentioned assumption that none of the data is cached in the L2 cache, our algorithm loads $M / \mathrm{TM} \times 16 / 1024 \times 64 \times 64 \times 2$ bytes from HBM to facilitate a matrix-matrix product of $2 \times 64 \times 64 \times M$ flops for an arithmetic intensity of $64 \times \mathrm{TM}$ flops per byte loaded from HBM. For our choice of $\mathrm{TM}=8$ our algorithm thus achieves an arithmetic intensity of 512 flops per byte loaded from HBM for the matrix-matrix product in every hidden layer, i.e., all layers except the input and output layers (where additional loads of the input from HBM or stores of the output to HBM are required). Together with a HBM bandwidth of approximately 2 TB/s (theoretical peak is $\sim 3.2 $ TB/s, 2 TB/s is a realistic bandwidth for complex workloads), this shows that the arithmetic intensity is high enough such that the available HBM bandwidth is not a limiting factor and that each layer is compute bound.

Analogously, the tiny-cuda-nn implementation presented in~\cite{muller2021real} loads the weight matrix for each layer $i$, $i=1, \ldots, \mathrm{nlayers}-1$, $M/128$ times from HBM (again discarding that the values may be cached in the L2 cache). This results in an arithmetic intensity of 128 flops per byte loaded from HBM. The Nvidia H100 PCIe GPU delivers a theoretical peak HBM bandwidth of approximately 2 TB/s~\cite{h100}. The theoretical peak throughput of the tiny-cuda-nn inference algorithm is thus reduced to 256 Tflops/s. Each layer in the CUDA implementation is therefore bound by the HBM bandwidth.

Extending the above model to consider the whole inference algorithm, including the loads of the input data and stores of the output, we get the following arithmetic intensities (also known as operational intensities; abbreviated as OI to prevent confusion with AI -- Artificial Intelligence) in dependence of the number of layers:
\begin{align*}
    \mathrm{OI}_{\text{SYCL}} = \frac{512 (\mathrm{nlayers}-1)}{16 + (\mathrm{nlayers}-1) } &,& \mathrm{OI}_{\text{CUDA}} = \frac{128 (\mathrm{nlayers}-1)}{4 + (\mathrm{nlayers}-1)}.
\end{align*}
For example, for a number of $\mathrm{nlayers}=6$, the arithmetic intensity of the SYCL implementation is 121.9 flops per byte, limiting the theoretical peak performance to approximately 243.8 Tflops/s.
The arithmetic intensity for the CUDA implementation is 71.1 flops per byte, resulting in a theoretical peak performance of approximately 142.2 Tflops/s.

Further extending the above model to the forward pass in the training step, where each $A_i$ is additionally stored in HBM, we get an arithmetic intensity of $512 (\mathrm{nlayers}-1) / (9\mathrm{nlayers}+7)$ flops per byte for the SYCL implementation ($\sim 49$ flops per byte for $\mathrm{nlayers}=6$; $\sim 98$ Tflops/s) and $128 (\mathrm{nlayers}-1) / (3\mathrm{nlayers}+1)$ for the CUDA implementation ($\sim 33.7$ flops per byte for $\mathrm{nlayers}=6$; $\sim 67.4$ Tflops/s) thus further reducing the theoretical peak performance for both codes.

Finally, considering the forward pass and the backward pass (loading the input, storing each $A_i$ and $D_i$ once, loading $B_i$ and $B_i^T$ according to the work-group size and the batch size, loading $A_i$ and $D_i$ once for the final calculation of the $G_i$) results in theoretical peak performance as displayed in Fig.~\ref{fig:theoretical_peak}.

Similar considerations as above for the accesses to SLM lead to an arithmetic intensity of 7.877 flops per byte accessed in SLM for the SYCL implementation and 25.6 flops per byte for the CUDA code. Thus, based on the theoretical peak SLM bandwidth of the targeted device, the SLM accesses limit the performance of our SYCL code to 393 Tflops/s (instead of 838 Tflops/s). This is never an issue for the training since the HBM bandwidth imposes a lower limit on the performance than the SLM bandwidth. Although, the SLM bandwidth limits the performance for the inference to at most to 393 Tflops/s when $\mathrm{nlayers} \geq 12$.

To emphasize the advantages of fused MLPs compared to non-fused MLPs, we compare the above arithmetic intensities to the arithmetic intensities of non-fused implementations in what follows. A non-fused implementation of the inference algorithm~\ref{alg:inference} would require for each layer a load of the matrices $A_i$ and $B_i$ with a subsequent store of $A_{i+1}$ to facilitate a single matrix-matrix product. Thus, the arithmetic intensity is bounded by 32 flops per byte loaded (independent of the number of layers) for the inference, which is up to 1/16-th of the arithmetic intensity of a fused implementation. Similar considerations for the training show that the fused training only increases the arithmetic intensity by up to a factor 2 at most (in the limit $\mathrm{nlayers} \to \infty$).

To summarize, Figure~\ref{fig:theoretical_peak} shows the theoretical peak performance based on the above roofline analysis for the CUDA implementation on Nvidia's H100 GPU and the SYCL implementation on Intel's Data Center GPU Max 1550. It shows that our algorithm improves the theoretical peak performance of the CUDA implementation~\cite{muller2021real} by increasing the arithmetic intensity for the HBM accesses and that decreasing the arithmetic intensity for SLM accesses does not have a negative impact. The performance which may be achieved in practice is discussed in the following Sec.~\ref{sec:results}.
\begin{figure}
    \centering
    \includegraphics[width=\linewidth]{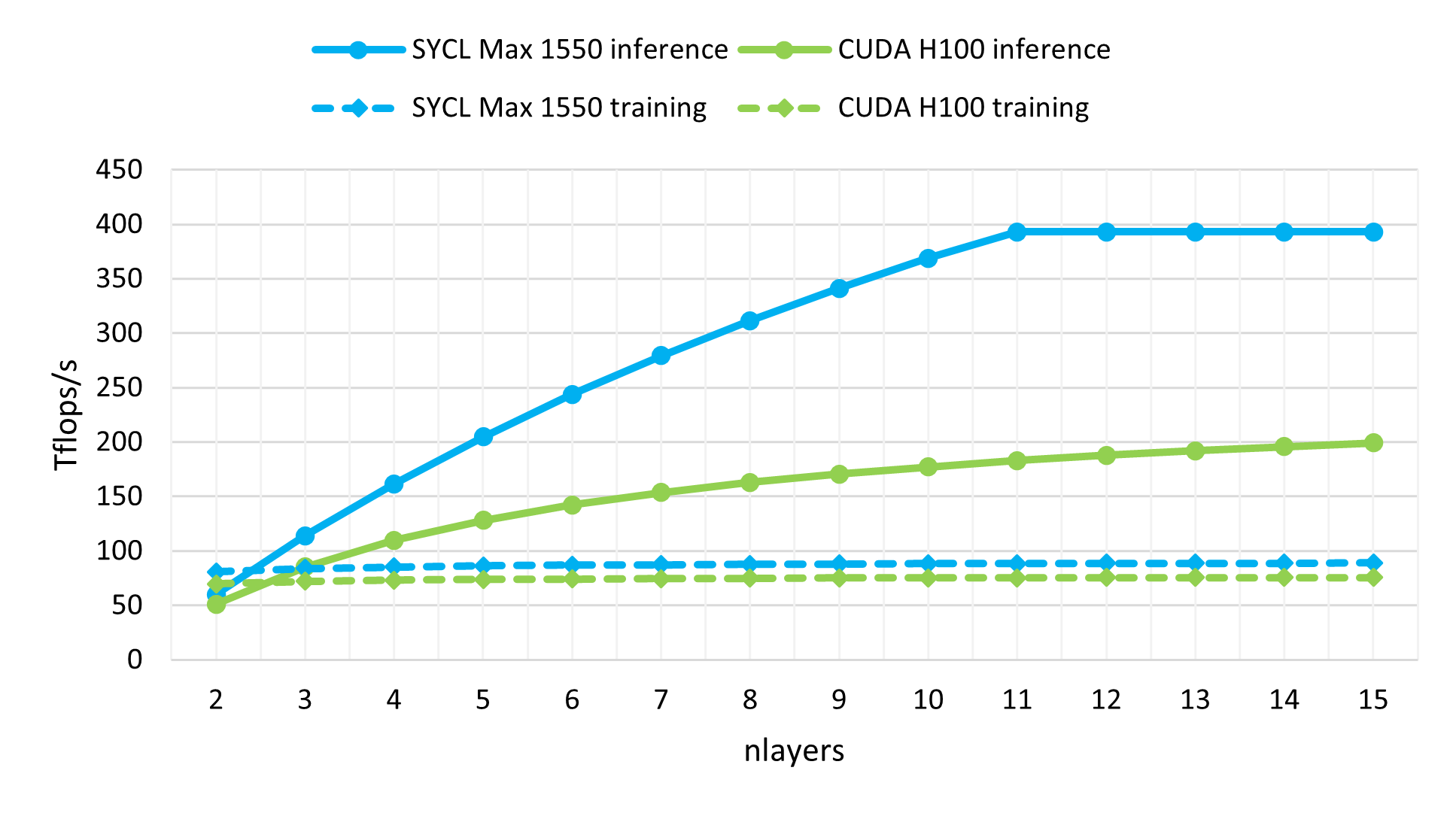}
    \caption{Comparison of the theoretical peak performance based on the the roofline analysis in Section~\ref{sec:roofline}.}
    \label{fig:theoretical_peak}
\end{figure}

\section{Results}
\label{sec:results}
In this section, we demonstrate how our fully-fused MLP implementation increases the performance of four commonly used AI tasks: non-linear function approximation, image compression, Neural Radiance Fields (NeRFs), and solving differential equations with PINNs (see Section~\ref{sec:apps}). For every comparison, we compare our SYCL implementation on an Intel Data Center GPU Max 1550 with the CUDA implementation~\cite{tiny-cuda-nn} on a Nvidia H100 GPU and PyTorch using both Intel Extension for PyTorch (IPEX) \cite{ipex} and CUDA backend. The results of the comparison can be seen in Table \ref{tab:comparison_all}. 
\begin{table}[h]
    \centering
    \caption{Training and Inference times for our implementation (SYCL), the fully-fused implementation from~\cite{muller2021real} (CUDA) and PyTorch using both IPEX and CUDA. The numbers in the brackets indicate the time relative to our implementation.}
    \label{tab:comparison_all}
\begin{tabular}{lll|l}
\multicolumn{1}{l}{}               &            & Training [s]	$\downarrow$ & Inference [s]	$\downarrow$ \\
                                   \hline
\multirow{4}{*}{Benchmark}         & SYCL       &    \textbf{0.58}     &         \textbf{0.083} \\
                                   & CUDA       &     0.79 (1.36)   &  0.236 (2.84)          \\
                                   & PyTorch (IPEX) &    4.57 (7.88)    &  2.178 (26.24)           \\
                                   & PyTorch (CUDA) &    4.32 (7.45)      &   1.543 (18.55)         \\
                                   \hline
\multirow{4}{*}{Image compr.} & SYCL       &   \textbf{9.20}      &  \textbf{2.041}        \\
                                   & CUDA       &     16.10 (1.75)    &     3.514 (1.72)        \\
                                   & PyTorch (IPEX) &   41.75 (4.54)  &       12.84 (6.29)   \\
                                   & PyTorch (CUDA) &      36.66 (3.98)    &    15.46 (7.57)         \\
                                   \hline
\multirow{4}{*}{NeRF}              & SYCL       &     \textbf{1.93}    &    \textbf{0.302}       \\
                                   & CUDA       &    2.05 (1.06)      &      0.477 (1.58)       \\
                                   & PyTorch (IPEX) & 15.09 (7.82)  &      9.050 (29.97) \\
                                   & PyTorch (CUDA) &     10.07 (5.21)     &          3.841 (12.72)   \\
                                   \hline
\multirow{4}{*}{PINNs}             & SYCL       &   \textbf{0.55}    &    \textbf{0.088}         \\
                                   & CUDA       &       0.60  (1.09)  &     0.108  (1.22)      \\
                                   & PyTorch (IPEX)    &   1.92 (3.49)     &  0.788 (8.95)     \\
                                   & PyTorch (CUDA) &  1.85 (3.36)         &  0.646 (7.34)           
\end{tabular}
\end{table}

To ensure a fair comparison of the forward and backward performances and to disregard differences in the implementation of the loss functions and optimiser, we only measure the time of the code where the fusion of the operators is relevant for, i.e., forward and backward calculation (cf. Algs.~\ref{alg:inference}, \ref{alg:training}).
The benchmark protocol is identical for both implementations: every forward and backward pass is called for the same amount of episodes, the same batch sizes, MLP architecture, and ReLU activation functions and a linear output activation function.

The CUDA implementation~\cite{tiny-cuda-nn} is tested on a dual-socket system with two Intel Xeon Platinum 8480+ CPUs, 512GB DRAM and Nvidia H100 GPUs with Ubuntu 22.04 LTS as operating system. The code was compiled with CUDA version 12.3.52 (nvcc version built on Sept. 8 19:17:24PDT 2023). The driver version is 535.129.03.

The SYCL code, on the other hand, is tested on a dual-socket system with two Intel Xeon Platinum 8480+ CPUs, 512GB DRAM and Intel Data Center GPU Max 1550 GPUs with Ubuntu 22.04.3 LTS as operating system. We utilized the mpiicpx compiler (Intel oneAPI DPC++/C++ Compiler 2023.2.0 (2023.2.0.20230721)) included in the oneAPI 2023.2 toolkit and an unreleased engineering driver\footnote{agama-ci-devel 682.16}. To scale our implementation to both Xe Stacks~\cite{xe_arch} in an Intel GPU, we use Intel MPI version 2021.10.0.

\subsection{Non-linear Function Approximation}
\label{sec:benchmark}
The goal of non-linear function approximation is to train a Neural Network $y = \hat{f}_{\theta}(x)$ to approximate a given non-linear function $f: \mathbb{R}^{K} \rightarrow \mathbb{R}^{N}$. During the training the residual loss between target function $f$ and approximation $\hat{f}_{\theta}$ is typically chosen as the mean square error, which is depicted in what follows for a batch size $M$:
\begin{equation}
    L(f,\hat{f}_{\theta},x_1,\ldots,x_M) = \sum_{i=1}^M \|f(x_i) - \hat{f}_{\theta}(x_i)\|^2.
\end{equation}

In this section, the performance of our SYCL implementation is compared to the CUDA implementation~\cite{tiny-cuda-nn} for various batch sizes $M$. To determine relevant batch sizes, we investigated the distribution of batch sizes used in MLP-based methods~\cite{tancik2022block, wu2023hyperinr, reiser2021kilonerf, sun2022direct, barron2021mip, hadadan2021neural, garbin2023stochastic, yu2022tensorf, yu2022tensorf, garbin2021fastnerf, fridovich2022plenoxels, yu2021plenoctrees, lu2019deeponet, wynn2023diffusionerf, uchytil2023function, kim2022neuralvdb, wang2022respecting, wang2021neus, raghavan2023neural, cuomo2022scientific, devkota2023efficient, lee2017estimating, sun2020model, saragadam2022miner, zhu20233d, mao2020ladabert, rebain2021derf, lindell2021autoint, neff2021donerf, neff2021donerf, sitzmann2021light, yu2023dylin, park2021nerfies, alkhalifahphysics, takamoto2023learning, he2023neural, cho2023separable, raissi2019physics, sankaran2022impact, sharma2022accelerated, wang2021understanding, biesek2023burgers} in the fields of NeRFs, Neural Compression, and PINNs mentioned in Appendix~\ref{sec:nerf_explanation}-~\ref{sec:piml_explanation}. The results of this investigation are summarized in Table~\ref{tab:batch_sizes}. The distribution appears to be approximately normal, with the majority of batch sizes falling between $2^{16}$ and $2^{22}$ suggesting that many MLP-based methods in these fields use large batch sizes, where our fully-fused implementation reaches maximal occupancy and performance (see Fig. \ref{fig:regression_result}).

\begin{table}[h]
\centering
\caption{Batch Sizes of commonly used MLP applications}
\label{tab:batch_sizes}
\begin{tabular}{l|llll}
Batch Size Interval & $[2^5, 2^{11})$ & [$2^{11}, 2^{15}]$ & $[2^{16},2^{22}]$ & $( 2^{22}, 2^{24}] $ \\\hline
\# Occurences       & 6          & 17              & 22              & 1         
\end{tabular}
\end{table}

For NeRFs, the batch size is the number of rays per batch and corresponding samples per ray during ray tracing. We the formula in \cite{muller2022instant} to calculate the batch size: number of rays per batch $\times$ number of samples per ray for the full method. 
For Neural Compression, and Partial Differential Equations (PDEs), since many methods in these representation learning use full-batch gradient descent as the training strategy, we adopt the dataset size as the full batch size for both the training and inference steps. 

Based on this data, batch sizes larger than $2^{22}$ elements are not commonly used and thus not considered in this test. Small batch sizes, in turn, are also removed from the test since the occupancy decreases proportional to the batch size for batch sizes smaller than $2^{16}$ (cf. Sec.~\ref{sec:implementation}). The performance is thus limited by the occupancy and fusing the operations does not provide any benefit. 

In our test we learn an MLP that approximates a non-linear function $f: \mathbb{R}^{64} \rightarrow \mathbb{R}^{64}$. We measure the throughput for our SYCL implementation on a single Xe Stack and both Xe Stacks of an Intel Data Center GPU Max 1550 and compare the results to the throughput measured with the CUDA implementation~\cite{tiny-cuda-nn} on a Nvidia H100 GPU. For both implementations, we use a network width of $64$ with input and output width of $64$ elements. As indicated above, the batch size $M$ varies from $2^{11}$ to $2^{22}$, and we run the benchmark for $\mathrm{niter}$ iterations calculated as $\mathrm{niter} = \text{max}(1000\cdot\frac{  2^{18}}{M} , 250)$. Further, we choose four hidden layers, i.e., $\mathrm{nlayers}= 6$.

\begin{figure}
    \centering
    \begin{subfigure}{\linewidth}
        \includegraphics[width=\linewidth]{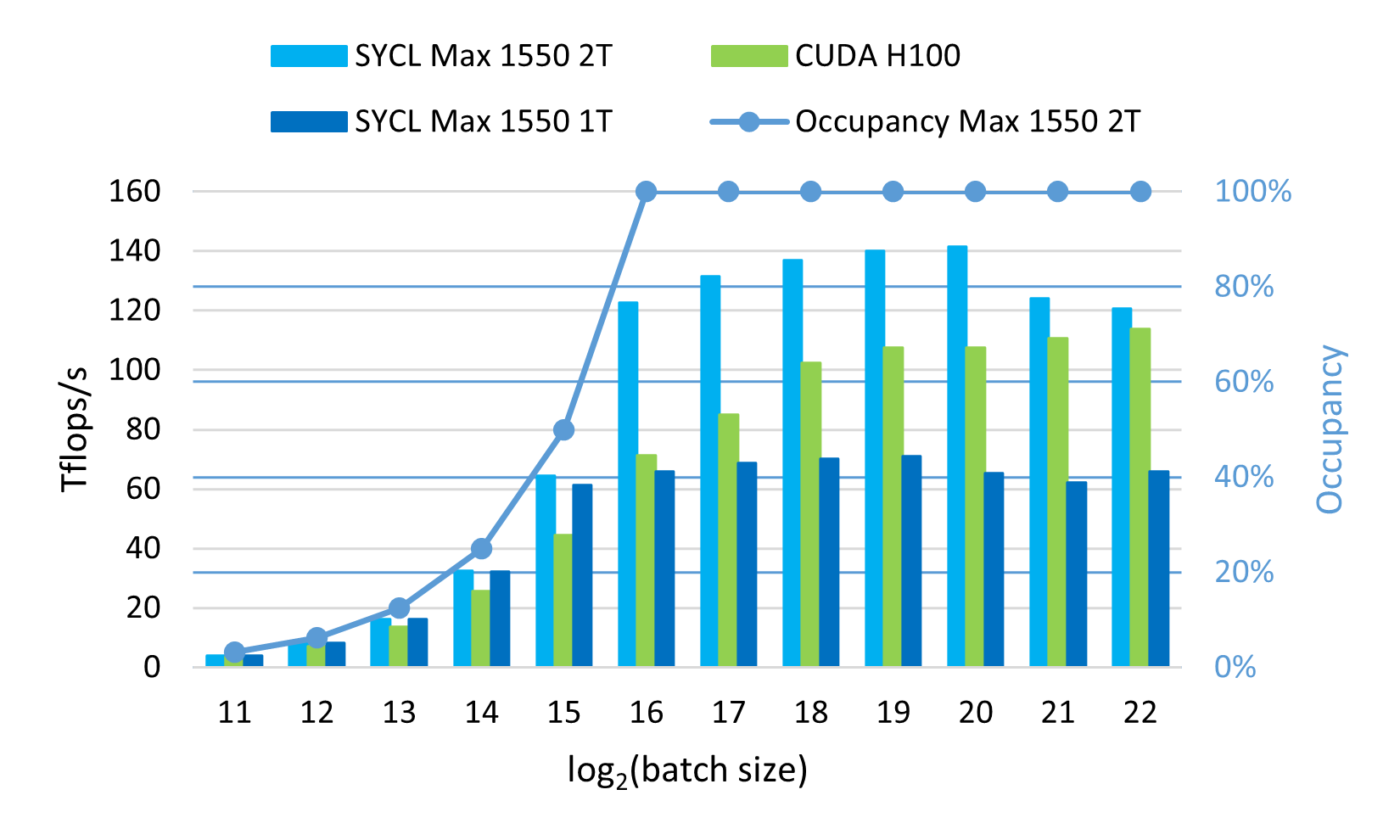}
    \end{subfigure}%
    
    \begin{subfigure}{\linewidth}
        \includegraphics[width=\linewidth]{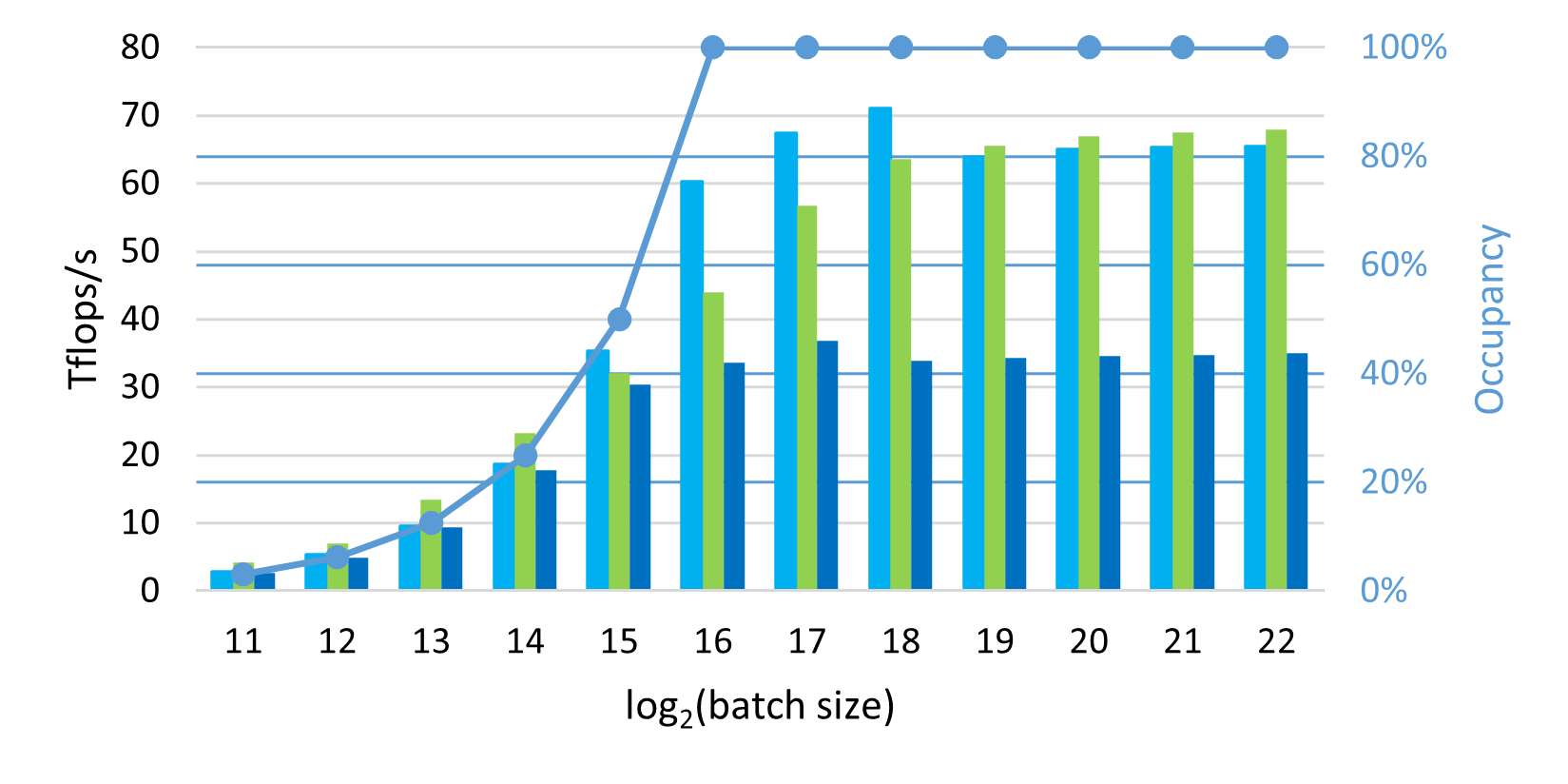}
    \end{subfigure}
    \caption{Performance of the inference (top) and training (bottom) of our SYCL implementation on a single tile of the Intel Data Center GPU Max 1550 (dark blue) and two tiles of the same Intel GPU (light blue) compared to the CUDA code on a Nvidia H100 GPU (green) for $\mathrm{nlayers} = 6$. The occupancy of the SYCL code on the two tiles of the Intel Data Center GPU is given as the blue line. The x-axis denotes the batch size (i.e., $M$) from $2^{11}$ inputs to $2^{22}$. }
    \label{fig:regression_result}
\end{figure}

The resulting performance graphs for training and inference can be seen in Figure~\ref{fig:regression_result} and the relative performance gain of our method compared to CUDA can be seen in Table \ref{tab:benchmark_result}. The figure shows that the performance in the training case (bottom graph of Fig.~\ref{fig:regression_result}) is similar for both implementations. This coincides well with the roofline predictions from Section~\ref{sec:roofline}. It is also interesting to note that the peak performance of approximately 70 Tflops/s is close to the roofline prediction of $\sim87 $ Tflops/s thus indicating that the problem is indeed bound by the HBM bandwidth.

\begin{table}[h]
    \centering
    \caption{Relative Performance Gain of SYCL compared to CUDA for Function Approximation}
    \begin{tabular}{c|ccc}
             & Mean  & Min & Max \\
    \hline
    Training throughput & 1.01 & 0.74 & 1.37\\
    Inference throughput & 1.34 & 1.03 & 1.68
    \end{tabular}
    \label{tab:benchmark_result}
\end{table}
The inference results (top graph of Fig.~\ref{fig:regression_result}), in contrast, show that the SYCL implementation outperforms the CUDA implementation. For the smaller batch sizes even when utilizing only half of the Intel device. This coincides again well with the roofline analysis in Section~\ref{sec:roofline}. Although, the measured peak performance of approximately 140 Tflop/s falls short of the predicted $\sim244$ Tflops/s. Our performance investigations have shown that this is due to scoreboard ID stalls occurring when loading the weights matrices from SLM thus indicating that the performance of our problem is not bound by the memory bandwidth but by the SLM memory latency. Investigation how to eliminate this latency are ongoing.

To fully show demonstrate the performance of our approach, we measure the time required to calculate 1000 iterations for non-linear function approximation using an MLP of 11 hidden layers and network width of 64. The batch size is chosen as $M = 2^{17}$. The results can be seen in the Benchmark row of Table \ref{tab:comparison_all}. Our SYCL implementation reaches performance gains of up to 7.88 times faster, and up to 26.24 times faster for training and inference respectively. 

As mentioned in Section~\ref{sec:implementation}, the Intel GPU is not fully occupied for batch sizes smaller than $2^{16}$ resulting in reduced performance. Note that maximum performances for the SYCL implementation are not achieved for the largest batch sizes, but for those batch sizes where the occupancy is maximal while the problem is sufficiently small to fit in the L2 cache.

\subsection{Image Compression}
For Image Compression, the MLP $\hat f_\theta$ learns the color distribution of an image represented as function $f: \mathbb{R}^{2} \rightarrow \mathbb{R}^{1}$ that maps from 2-D coordinates (pixels) to corresponding color. We use a network width of $64$. As NNs poorly learn high-frequent details of the target function $f$, encodings are typically used to mitigate this issue. Encodings $g: \mathbb{R}^{2} \rightarrow \mathbb{R}^{N}$ pre-process the 2D pixel input by directly mapping them into a spectrum of high-frequent signals, which are then fed into the MLP. The Multiresolution hash encoding $g: \mathbb{R}^{2} \rightarrow \mathbb{R}^{32}$ \cite{muller2022instant} is used, which can be seen as the input to the MLP. The final function approximator is thus $h(x) = \hat{f}(g(x))$, with Multiresolution Hash Encoding $g$ and MLP $\hat f: \mathbb{R}^{32} \rightarrow \mathbb{R}^{1}$.

The learning progress can be seen in Figure \ref{fig:einstein}. We used a network width of $64$ with input $32$, output width $1$, and batch size $K = 2304 \times 3072 = 7077888$ for 1000 iterations. After 10 steps the colors are not entirely correct, but the silhouette is recognisable. Then, over the next 900 steps, the colors and finer details are becoming gradually clearer until the training progress converges after 1000 steps. The whole training takes $9.2$s for the SYCL implementation and is 1.75 faster than the CUDA implementation and ~4 times faster than PyTorch. The image can be reconstructed (inference) within $2.41$s for SYCL (see row Image compr. in Table \ref{tab:comparison_all}).

For comparison, storing the image at a resolution of $2304 \times 3072$ with greyscale values costs 2.3MB for a JPEG image, while storing 12288 half precision weights costs $25$ kb.

\begin{figure}[h]
  \centering
  \includegraphics[width=0.48\textwidth, page=1, trim=0 5cm 0 5cm, clip]{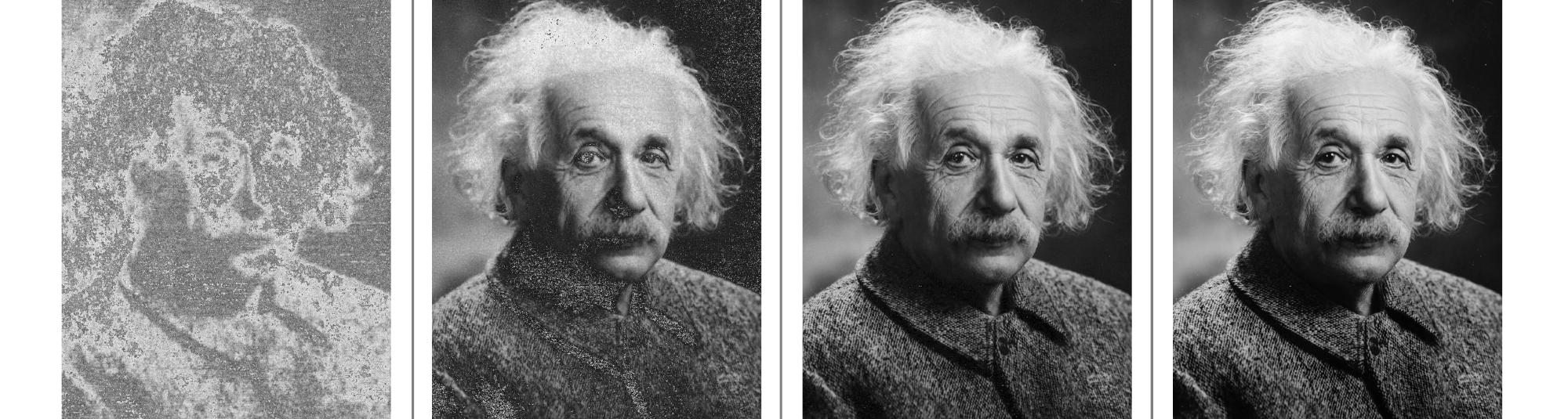}
  \caption{Training progress of Image Compression. The training converges after 1000 steps. The training and inference process is performed for both the SYCL and CUDA implementation and the visualised progress if implemented as per \cite{tiny-cuda-nn}.}
  \label{fig:einstein}
\end{figure}

\subsection{Neural Radiance Fields (NeRF)}

For NeRFs, the goal is to learn a Radiance Field (see Appendix~\ref{sec:nerf_explanation}) $f: \mathbb{R}^{5} \rightarrow \mathbb{R}^{4}$. Similar to Image Compression, NeRFs require an encoding as input to the MLP for better representations of the high-frequent details of the object. Multiresolution hash encoding is used for this task resulting in a 32-dimensional feature vector as input to the MLP. A network width of $64$ is chosen with input width $32$, output width $4$, and a batch size $K = 1048576$. For details about the implementation of the NeRF algorithm, please see~\cite{cai2023clnerf}.

The training of the MLP without encoding and volume rendering is conducted within $1.93$s for training and $0.302$s inference in our SYCL implementation. Compared to the CUDA implementation, this is 1.06 and 1.58 times faster for training and inference respectively. The generated images can be seen in Figure~\ref{fig:clnerf_results}.

\begin{figure*}[h]
  \centering
  \includegraphics[width=0.8\textwidth, page=1, trim=1cm 0cm 1cm 0cm, clip]{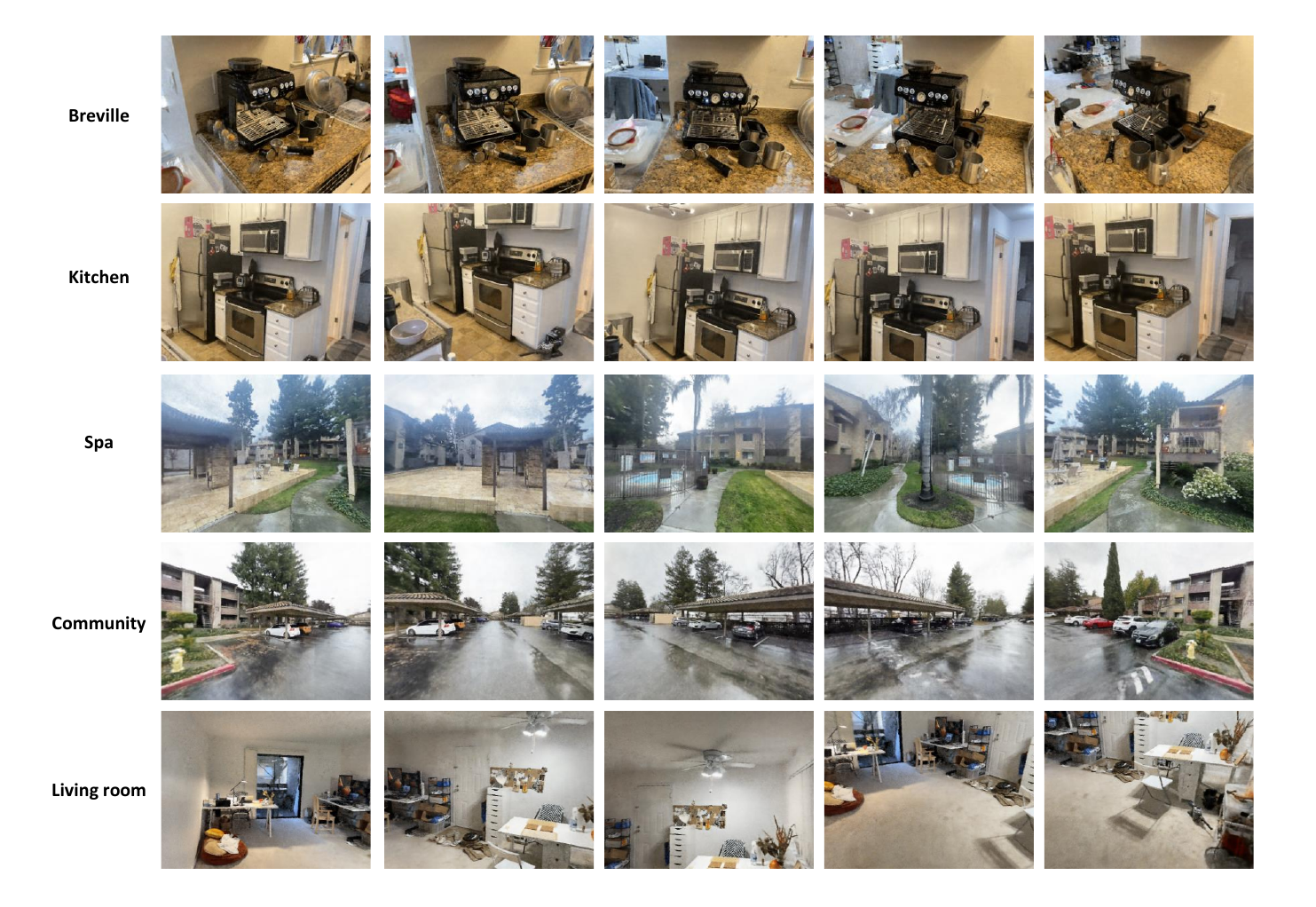}
  \caption{CLNeRF results on 5 scenes for breville, kitchen, spa, community, and living room. Each row represents a different scene and each column shows the scene rendering results from different views using the MLP as NeRF. }
  \label{fig:clnerf_results}
\end{figure*}

\subsection{Physics-Informed Neural Networks (PINN)}

\begin{figure*}[h]
  \centering
  \includegraphics[width=\textwidth, page=1, trim=0cm 10cm 0cm 0cm, clip]{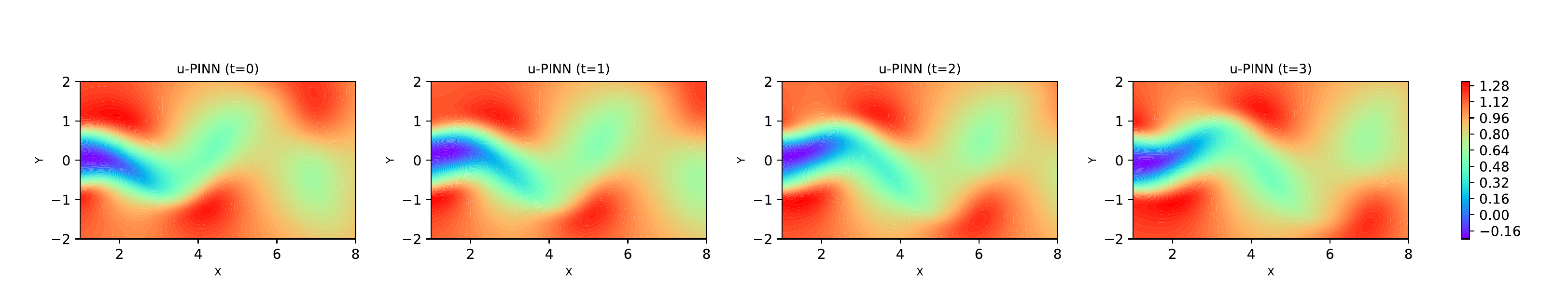}
  \caption{Visualisation of learned solution to Navier Stokes Equation. Mean Absolute Percentage Error: 0.1\%. Results generated using DeepXDE \cite{lu2021deepxde}.}
  \label{fig:pinns_results}
\end{figure*}
To demonstrate how our method can be applied to solve Partial Differential Equations (PDE), we leverage the PINNs framework \cite{raissi2019physics} to train an MLP that represents the solution of the following two-dimensional Navier-Stokes Equation:
\begin{align}
    u_t + \lambda_1(uu_x + vu_y)  +p_x-\lambda_2(u_{xx} + u_{yy})&= 0 ,\label{eq:navier_stokes}\\
    v_t + \lambda_1(uv_x + vv_y)  +p_y-\lambda_2(v_{xx} + v_{yy})&= 0,\\
    u_x + v_y &= 0,
\end{align}
with velocity fields  $u(t,x,y)$  and $v(t,x,y)$ for the $x$ and $y$ components respectively,  pressure $p(t,x,y)$, and unknown parameters $\lambda = (\lambda_1, \lambda_2)$. For more details of the problem, please refer to Section 4.1.1 in \cite{raissi2019physics}.

We used DeepXDE \cite{lu2021deepxde} to find the PINNs $f(t,x,y)$, $g(t,x,y)$ that represent the solution of the Navier stokes equations (\ref{eq:navier_stokes}). For details, see Appendix~\ref{sec:piml_explanation}.

The dataset and thus batch size consists of $M = 2^{17}$ data points. For 1000 iterations, the training time takes $0.55$s for SYCL, $0.60$s for CUDA yielding a performance gain of up to 3.49 compared to PyTorch. The evaluation of the PDE solution on $2^{17}$ collocation points requires $0.088$, which results in an improvement of up to 8.95x.

\section{Conclusion and Future Work}
In this paper, we presented a SYCL implementation of fully-fused Multi-Layer Perceptrons (MLPs) on Intel Data Center GPU Max. Our approach focuses on maximizing data reuse within the general register file and the shared local memory, minimizing the need for slower global memory accesses. This results in a significant increase in the arithmetic intensity, which leads to improved performance, especially for inference.

Our implementation outperforms an equivalent CUDA implementation for MLPs with width 64 by a factor of up to 2.84 in inference and 1.75 in training in our tests, demonstrating the effectiveness of our approach, and outperforms the PyTorch implementation by up to a factor of 30. We further showcased the efficiency of our implementation in three significant areas: Image Compression, Neural Radiance Fields (NeRF), and Physics-Informed Machine Learning. Across all these domains, our approach demonstrated substantial improvements, achieving factors up to 30 times when compared to conventional PyTorch implementations and up to 2.84 times over highly optimized CUDA implementations. 

In the future, we aim to further optimize our implementation. A strong focus will be a more efficient usage of registers which may enable further prefetching of the weights matrices, thus reducing the stalls. In addition, we may be able to reduce the utilization of SLM and enable the loads of multiple weights matrices in the SLM, which would reduce the number of necessary barriers. Another important aspect will be to increase the occupancy for small batch sizes by either increasing the arithmetic intensity for smaller values of $\mathrm{TM}$ or by partitioning the work along the $K$ dimension, which requires the usage of atomic operations. Lastly, an important optimization would be the fusion of the final matrix products $A_i^T D_i$ (cf. Alg.~\ref{alg:training}) into the backward pass. It could potentially increase the performance of the training by up to a factor 2, based on the roofline model.

Our implementation is open-sourced to allow for wider usage and contributions from the community. The code can be found at \url{https://github.com/intel/tiny-dpcpp-nn}.

In addition to further performance optimization, we also plan to explore the use of Intel's ESIMD SYCL extension for our implementation and compare it to our existing SYCL implementation. Our past experience with ESIMD showed that it enables a better control over the register usage and exposes cache controls for load and store operations. 

In future work, we aim to generalize our library to various data types and larger network widths. In addition, we plan to implement an optimized version for Intel's Arc GPUs.

\section*{Acknowledgment}
We'd like to thank Jia Xu and Jing Xu for their help and advise during the development of tiny-dpcpp-nn.

\section*{Disclaimers}
Performance varies by use, configuration and other factors. Learn more on the Performance Index site. 

Performance results are based on testing as of dates shown and may not reflect all publicly available updates. No product or component can be absolutely secure. 

Your costs and results may vary. 

Intel technologies may require enabled hardware, software or service activation.

\copyright Intel Corporation. Intel, the Intel logo, and other Intel marks are trademarks of Intel Corporation or its subsidiaries. Other names and brands may be claimed as the property of others.

\bibliographystyle{IEEEtran}
\bibliography{reference}

\newpage
\appendix
\label{sec:appendix}
\subsection{Neural Radiance Fields}
\label{sec:nerf_explanation}
Recently, MLPs have been instrumental in the emerging Neural Rendering (NeRFs) field~\cite{mildenhall2021nerf}. NeRF uses \textbf{MLPs} to represent the scene as a continuous function $f_\theta (x,y,z,\theta,\phi) = (r,g,b, \sigma)$, which is parameterized by a MLP network that takes a 3D point $(x,y,z)$ and a viewing direction $(\theta,\phi)$ as input and outputs a 4D vector $(r,g,b,\sigma)$ where $(r,g,b)$ is the color and $\sigma$ is the density. 

To render an image from a given camera pose, we use ray casting to sample points along each ray and compute the accumulated color and density using volume rendering by the following function:
\begin{align}
    \label{color render}
     \mathbf{{C(r)}}=\int^{t_{far}}_{t_{near}} T(t) \sigma(r(t)) c(r(t),d) dt 
\end{align}

where $C(r)$ is the final color of the ray $r$, $T(t) = \exp(-\int_{t_{near}}^{t_{far}} \sigma(t') dt')$ is check the occlusion through the ray traveling from the near plane $t_{near}$ to the far plane $t_{far}$ and $r(t) = o + td$ is the ray equation with origin $o$ and direction $d$.

To train the MLP, NeRF uses a combination of coarse and fine rendering losses. The reconstruction loss measures the difference between the rendered color and the ground-truth color for each pixel in the training images. The total loss is defined as:
\begin{equation}
    L = \sum_{r\in R} \|\hat{C}_c(r) - C_{gt}(r)\|_2^2 + \|\hat{C}_f(r) - C_{gt}(r)\|_2^2
\end{equation}

where $r$ is the ray that passes through the pixel, $C_{gt}(r)$ is the ground truth color of the pixel. By minimizing this loss, the MLP learns to represent the scene as a neural radiance field that can synthesize novel views from any camera pose.

\subsection{Neural Compression}
\label{sec:neural_compression_explanation}
There has also been an increased focus on using MLPs for Neural Compression, where MLPs aim to learn non-linear mapping as a function approximation to compress data and preserve the most relevant information. To this end, the MLP can be used as the encoder and/or the decoder in the compression pipeline. The encoder transforms the original data into a compressed representation, and the decoder aims to reconstruct the data. MLPs can be optimized to minimize the difference between the original and the reconstructed image. The loss function for neural data compression can be defined as:
\begin{equation}
    L = \lambda D(x, \hat{x}) + R(z)
\end{equation}

where $x$ is the original image, $ \hat{x}$ is the reconstructed image, $z$ is the representation, $D$ is a distortion measure such as mean squared error, $R$ is a rate measure such as entropy, and $\lambda$ is a trade-off parameter that balances the distortion and the rate.

\subsection{Partial Differential Equations (PDEs)}
\label{sec:piml_explanation}

Partial Differential Equations (PDEs) have attracted much attention recently. However, solving PDEs analytically is often impossible or impractical, and numerical methods can be costly or inaccurate.

MLPs can be used to solve PDEs in a data-driven and physics-informed way. One of the approaches is to use Physics-Informed Neural Networks (PINNs), which leverage MLPs to learn and approximate the solutions to complex physical systems. Given the underlying PDE and initial, boundary conditions embedded in a loss function, a coordinate-based neural network is trained to approximate the desired solution. PINNs take the form of:
\begin{equation}
    u(x,t)=f_\theta(x,t)
\end{equation}
where $u(x,t)$ is the unknown solution, $f_\theta(x,t)$ is an MLP with parameters $\theta$, $x$ is the spatial variable, and $t$ is the temporal variable. The MLP is trained to satisfy the boundary and initial conditions of the PDE, as well as the PDE itself. The loss function can be defined as:
\begin{equation}
    L = \sum_{i=1}^N \|u(x_i, t_i) - f_{\theta}(x_i, t_i)\|^2 + \sum_{j=1}^M \|F(x_j, t_j)\|^2
\end{equation}
where $N$ is the number of data points, $M$ is the number of collocation points, $u(x_i, t_i)$ is the observed or prescribed value of the solution at $(x_i, t_i)$, $f_\theta(x_i, t_i)$ is the predicted value of the solution at $(x_i, t_i)$, and $F(x_j,t_j) $is the residual of the PDE at $(x_j,t_j)$, which is computed by applying automatic differentiation to the MLP.
By minimizing this loss, the MLP learns to approximate the solution of the PDE that is consistent with the data and the physics.

\end{document}